\RequirePackage{fix-cm}
\documentclass[twocolumn]{svjour3}          
\smartqed  

\usepackage{times}
\usepackage{epsfig}
\usepackage{graphicx}
\usepackage{amsmath}
\usepackage{amssymb}
\usepackage{dsfont}
\usepackage{bbding}
\usepackage{url}
\usepackage{float}
\usepackage{subfig}
\usepackage{bm}
\usepackage{pifont}
\usepackage{booktabs}
\usepackage{color,xcolor}
\usepackage{tcolorbox}
\usepackage{colortbl}
\usepackage{algorithm}%
\usepackage{algorithmic}%
\usepackage{array}
\usepackage{wrapfig}
\usepackage{multirow}


\begin{document}
	
	\title{Creatively Upscaling Images with Global-Regional Priors
	}
	

	\author{Yurui Qian \and Qi Cai \and Yingwei Pan \and Ting Yao~\textmd{\Envelope} \and Tao Mei
	}
	
	\authorrunning{Yurui Qian, Qi Cai, Yingwei Pan, Ting Yao, Tao Mei} 
	
	\institute{Yurui Qian$^{1}$ \at \email{qyr123@mail.ustc.edu.cn}          
		\and
		Qi Cai$^{2}$ \at \email{cqcaiqi@hidream.ai}
		\and
		Yingwei Pan$^{2}$ \at  \email{pandy@hidream.ai}
		\and
		Ting Yao$^{2}$ (Corresponding author) \at \email{tiyao@hidream.ai}
		\and
		Tao Mei$^{2}$ \at \email{tmei@hidream.ai}
		\and
		$^1$\;\; University of Science and Technology of China \\
		$^2$\;\; HiDream.ai Inc. 
	}
	
	\date{Received: date / Accepted: date}

	\maketitle
	
	\begin{abstract}
		Contemporary diffusion models show remarkable capability in text-to-image generation, while still being limited to restricted resolutions (e.g., 1,024$\times$1,024). Recent advances enable tuning-free higher-resolution image generation by recycling pre-trained diffusion models and extending them via regional denoising or dilated sampling/convolutions. However, these models struggle to simultaneously preserve global semantic structure and produce creative regional details in higher-resolution images. To address this, we present C-Upscale, a new recipe of tuning-free image upscaling that pivots on global-regional priors derived from given global prompt and estimated regional prompts via Multimodal LLM. Technically, the low-frequency component of low-resolution image is recognized as global structure prior to encourage global semantic consistency in high-resolution generation. Next, we perform regional attention control to screen cross-attention between global prompt and each region during regional denoising, leading to regional attention prior that alleviates object repetition issue. The estimated regional prompts containing rich descriptive details further act as regional semantic prior to fuel the creativity of regional detail generation. Both quantitative and qualitative evaluations demonstrate that our C-Upscale manages to generate ultra-high-resolution images (e.g., 4,096$\times$4,096 and 8,192$\times$8,192) with higher visual fidelity and more creative regional details.
	\end{abstract}

	\section{Introduction}
	In recent years, diffusion models \cite{ddpm,ldm,ddim} emerge as a new trend of powerful generative architecture in various image synthesis tasks. With the availability of web-scale text-image paired data, several practical diffusion-based systems (e.g., DALL-E 3 \cite{betker2023improving}, Imagen \cite{saharia2022photorealistic}, SDXL \cite{sdxl}) have been developed to synthesize high-quality images with remarkable visuals. However, these systems are mostly limited to generating images of relatively low resolution (e.g., 1,024$\times$1,024). This severely hinders their applicability in many real-world scenarios that require higher-resolution images, such as electronic advertising and film production.
	
	\begin{figure*}[tb]
		\centering
		\includegraphics[width=1.95\columnwidth]{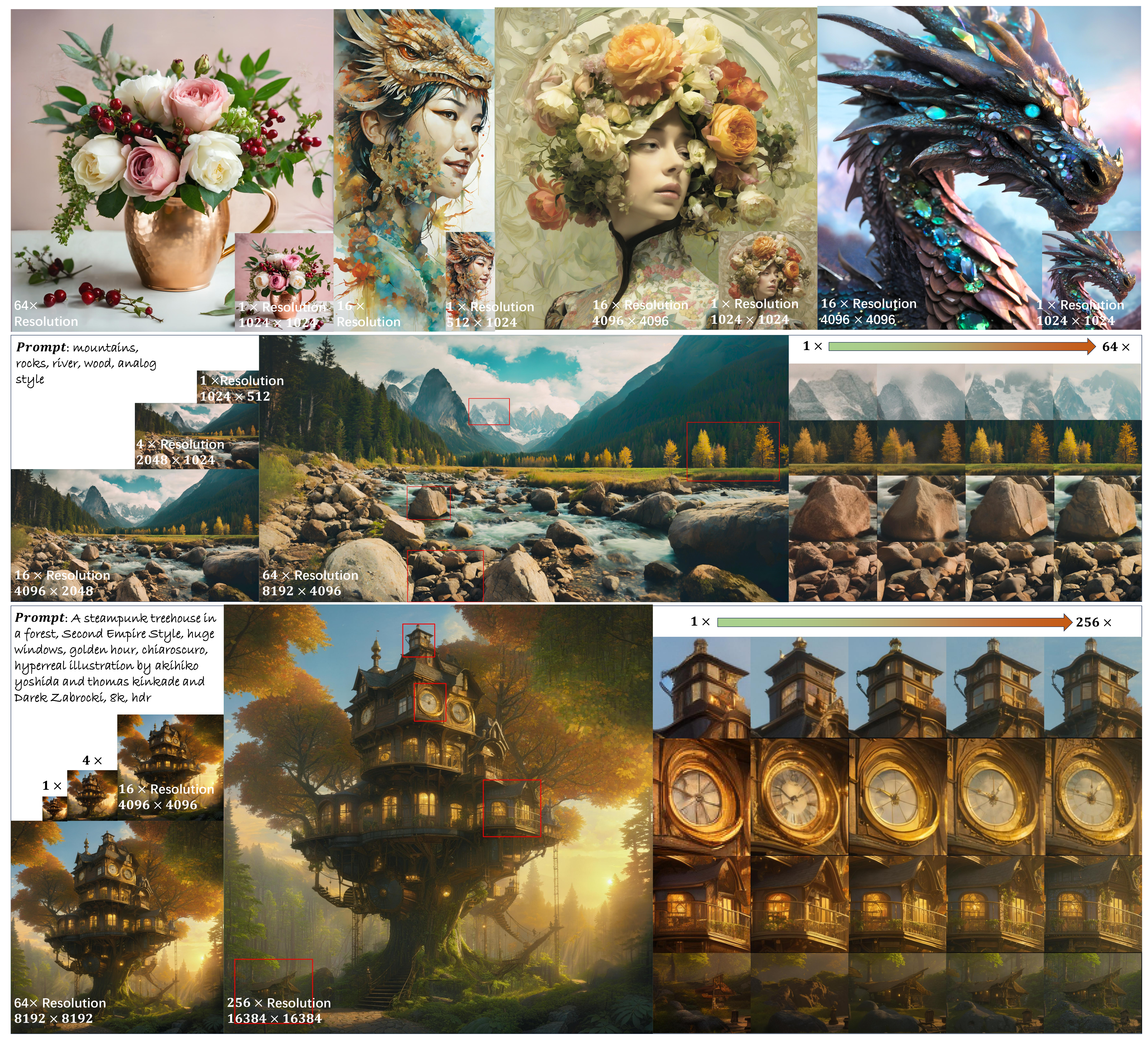}
		\caption{
			C-Upscale for high resolution image generation. SDXL generates images with resolutions up to $1,024^2$ and C-Upscale subsequently upscales images at 4$\times$, 16$\times$, and even 256$\times$. The regions in red boxes are shown in zoom-in view (right side).
		}
		\label{fig:intro}
	\end{figure*}
	The general objective of higher-resolution image generation is three-fold: 1) \textbf{visual quality}: the synthesized higher-resolution image should be of high-quality with few artifacts; 2) \textbf{global semantic/structure alignment}: the synthesized higher-resolution image should conform to the given global prompt or the global structure of pre-made low-resolution image derived from global prompt; 3) \textbf{regional creativity}: the synthesized higher-resolution image is expected to exhibit more creative and reasonable regional details, and this unique point also differentiates this task from typical image super-resolution \cite{dong2023deep,stablesr,realesr,xin2023advanced} who strictly adhere to the input low-resolution images and result in regional smoothing instead of creating details. One straightforward way is to directly apply pre-trained diffusion model to produce higher-resolution images, but it will inevitably result in several issues like object distortion and repetition \cite{demofusion,scalecrafter}. It is also resourcefully expensive to re-train diffusion models tailored for higher-resolution images, and the deployment of such diffusion models at inference requires a quadratic memory footprint.
	
	To alleviate these limitations, recent pioneering practices start to remould off-the-shelf text-to-image diffusion models with regional denoising \cite{multidiffusion} or dilated sampling/convolutions \cite{demofusion,scalecrafter} to enable higher-resolution image generation in a tuning-free fashion. For example, MultiDiffusion \cite{multidiffusion} highlights the potential of combining overlapping regional denoising paths for generating high-resolution images (e.g., panoramic images). Nevertheless, each regional denoising path is merely driven by the given global prompt, and such global-regional semantic discrepancy will result in object repetition issue. DemoFusion \cite{demofusion} and ScaleCrafter \cite{scalecrafter} further introduce several strategies (e.g., progressive upscaling, skip residual, and dilated sampling/convolutions) to enhance global semantic/structure alignment against global prompt/pre-made low-resolution image or enlarge receptive field. This direction reduces the regional repetitive content, while leaving the regional creativity in higher-resolution image generation under-explored.
	
	To explore the feasibility of jointly triggering global semantic/structure alignment and regional creativity, our work shapes a new tuning-free image upscaling paradigm (namely C-Upscale) on the basis of amplified guidance of both global and regional semantics. Our launching point is to excavate three kinds of prior knowledge (i.e., global structure prior, regional attention prior, and regional semantic prior) from global prompt and estimated regional prompts, and facilitate higher-resolution image generation with these global-regional priors. Technically, C-Upscale initially generates the low-resolution image from global prompt via pre-trained diffusion model that conveys global structure information. After employing low-pass filters over this low-resolution image, the learnt low-frequency component basically reflects global scene structure and we take it as global structure prior to guide higher-resolution image generation, thereby pursuing global structure alignment. Next, to further alleviate the global-regional semantic discrepancy, we filter out the semantically irrelevant cross-attention between global prompt and each region. As such, C-Upscale only retains the semantically relevant cross-attention in between as the regional attention prior to strengthen regional denoising, thereby mitigating object repetition issue. Meanwhile, we capitalize on pre-trained Multimodal LLM \cite{liu2024llavanext} to produce regional prompts that depict rich descriptive details of each region. C-Upscale additionally takes the estimated regional prompts as regional semantic prior to amplify the creativity of regional detail generation.
	
	In summary, we have made the following contributions: 1) C-Upscale is shown capable of leveraging global-regional priors to faithfully synthesize
	higher-resolution images with better global semantic alignment and regional creativity. 2) The exquisitely designed C-Upscale is shown able to be seamlessly integrated into existing pre-trained diffusion models. 3) C-Upscale has been properly analyzed and verified through extensive experiments for upscaling both real-world and synthetic images to validate its efficacy.
	
	\section{Related Work}
	\noindent \textbf{Diffusion Models for Image Generation.}
	Diffusion models \cite{dhariwal2021diffusion,ddpm,ddim}, a powerful class of generative models, have emerged as a remarkable approach to producing high-quality images by learning complex data distributions. 
	Latent diffusion models (LDMs) \cite{ldm} reformulate the diffusion process into the latent space, demonstrating their effectiveness in generating visually compelling images. Subsequent research efforts, including SDXL \cite{sdxl}, Emu \cite{dai2023emu}, Dall-E 3 \cite{betker2023improving}, and others \cite{chen2023controlstyle,pixart,li2024playground,qian2024boosting,saharia2022photorealistic,zhang2024open,zhu2024sd}, have significantly advanced the field through improvements in network architectures, training data, and conditioning mechanisms.
	\cite{yang2024mastering} generates images with better text-to-image compositionality through harnessing chain-of-thought reasoning ability of Multimodal LLM. Diffusion models have also been extended to tackle image inpainting \cite{chen2024improving,quan2024deep,wan2024improving} and image-to-image translation tasks \cite{choi2021ilvr,fei2023generative}. A notable limitation of most existing diffusion models is that they are constrained by the resolution they were trained on, typically around one million pixels and not directly applicable to higher-resolution image generation tasks. C-Upscale falls within the domain of image-to-image translation and has the ability to produce higher-resolution images than those in the training stage.

	\noindent \textbf{High-Resolution Image Generation.}
	The challenge of generating high-resolution images has been a topic of significant interest in the field. Super-resolution models \cite{dong2015image,stablesr,realesr,zhang2021designing} upscale images to higher resolutions, closely adhering to the input low-resolution images. This strict adherence often results in regional smoothing and a lack of fine details typically present in high-resolution images. To overcome this limitation, researchers have recently turned to diffusion models for image upscaling. These models can be broadly categorized into tuning-based approaches and tuning-free approaches.	Tuning-based approaches \cite{guo2024make,ren2024ultrapixel,zheng2023any} involve designing specific network architectures and retraining the diffusion models. However, these methods are constrained by particular base model architectures and cannot be easily generalized to newly trained models. 
	
	Tuning-free approaches aim to adapt pre-trained diffusion models for generating higher-resolution images without additional training. The pioneering work, MultiDiffusion \cite{multidiffusion}, proposes fusing multiple diffusion paths to generate high-resolution and complex prompt-conditioned images. Subsequent approaches, including DemoFusion \cite{demofusion}, ScaleCrafter \cite{scalecrafter}, and FouriScale \cite{huang2024fouriscale}, incorporate dilated sampling/convolutions to mitigate object repetition issues commonly encountered in high-resolution image generation. Concurrent to our work, more studies have been proposed to address the object repetition problem. For instance, DiffuseHigh \cite{kim2024diffusehigh} directly replaces the low-frequency component of the high-resolution image with that from the low-resolution image, which may introduce inconsistencies between different frequencies. Moreover, ElasticDiffusion \cite{hajiali2023elasticdiffusion} employs gradients derived from the low-resolution image to maintain consistency between the upscaled image and the low-resolution image. However, this method is susceptible to high-frequency artifacts present in low-resolution images. In contrast, our proposed C-Upscale method utilizes the low-frequency components from low-resolution image to guide synthesis, effectively avoiding high-frequency artifacts such as blurriness and distortions while preserving the original structures. Alternatively, AccDiffusion \cite{lin2024accdiffusion} tackles the object repetition issue by deriving object masks through filtering the cross-attention scores of low-resolution image and masking the word tokens accordingly. While promising, AccDiffusion necessitates a specialized thresholding operation, which results in the loss of information captured by the attention scores. In comparison, our approach leverages cropped regional attention scores, which better preserves regional information.
	
	\section{Preliminaries}
	\label{sec:prelim}
	\textbf{Diffusion Models} for image generation are designed based on the principle of simulating a physical diffusion process to progressively transform Gaussian noises into high-fidelity images. The typical large-scale text-conditioned diffusion models are latent diffusion models \cite{ldm}, such as SDXL \cite{sdxl} and Emu \cite{dai2023emu}. Without loss of generality, we illustrate the diffusion process using SDXL as an example. It consists of a variational autoencoder (VAE) \cite{esser2021taming} that transforms the image $\mathbf{x} \in \mathbb{R}^{3 \times H \times W}$ into latent space $\mathbf{z}_0 \in \mathbb{R}^{4 \times H/8 \times W/8}$. In the forward process, the original sample $\mathbf{z}_0$ is progressively corrupted through the addition of noise. The noising process can be formulated as follows:
	\begin{align}
		&q(\mathbf{z}_t | \mathbf{z}_{t-1}) = \mathcal{N}(\mathbf{z}_t; \sqrt{\alpha_t}\mathbf{z}_{t-1}, ( 1- \alpha_t)\mathbf{I}), t = 1, \dots, T, \\
		&\mathbf{z}_{t} = \sqrt{\bar{\alpha}_t} \mathbf{z}_0 + \sqrt{1 - \bar{\alpha}_t} \boldsymbol{\epsilon}, \boldsymbol{\epsilon} \sim \mathcal{N}(0, I), \label{eq:x0_pred}
	\end{align}
	where $\alpha_t$ represents the predefined schedule governing the denoising process and $\overline{\alpha}_t = \prod_{i=1}^{t} \alpha_i$. To sample images with the inverse process, a random noise $\mathbf{z}_T$ is drawn from a standard normal distribution and progressively denoised with a denoising network $\boldsymbol{\epsilon}_{\boldsymbol{\theta}}(\cdot)$. The denoising process can be formulated as follows:
	\begin{equation} 
		\mathbf{z}_{t-1} = \boldsymbol{\epsilon}_{\boldsymbol{\theta}}(\mathbf{z}_t, t, p), \ t = 1, \dots, T.
	\end{equation}
	$\mathbf{z}_t$ represents the latent vector at denoising step $t$, and $p$ denotes the textual prompt utilized for conditioning the generation process. The denoising function $\boldsymbol{\epsilon}_{\boldsymbol{\theta}}$, parameterized by $\boldsymbol{\theta}$, typically corresponds to a UNet architecture \cite{ronneberger2015u}.
	
	\textbf{Diffuse-then-Denoise} \cite{meng2022sdedit} is a widely adopted technique in diffusion models for image-to-image translation. Starting with a source input image, this approach adds an intermediate level of noise (\textbf{diffuse}) to the input according to Eq. \ref{eq:x0_pred}, where $t$ specifies the noise level. The resulting noisy sample, $\mathbf{z}_t$, is then processed by the diffusion model, which iteratively \textbf{denoises} it to generate the output image. The value of $t$ plays a crucial role in balancing faithfulness to the input image and creativity stimulation in the output image. Specifically, smaller values of $t$ yield outputs closely resembling the input image, while larger values allow for more creativity. Notably, this strategy is computationally efficient, requiring only $t/T$ of the time required by a full-length diffusion generation process, where $T$ denotes the number of sampling timesteps.

\begin{figure}[h]
		\centering
		\includegraphics[width=0.89\columnwidth]{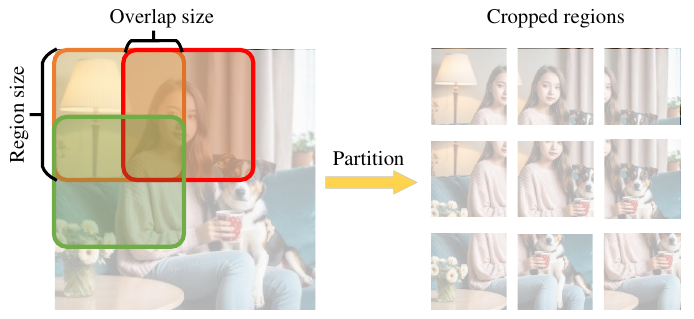}
		\caption{Visualization of the partitioning process, which divides the latent space into smaller overlapped cropped regions. The region size specifies the dimensions of each cropped region, while the overlap size defines the extent of overlap between adjacent regions.}
		\label{fig:crop}
	\end{figure}

	\textbf{Regional Denoising} \cite{multidiffusion,demofusion} adopts a ``divide-and-conquer'' strategy to process high-resolution images exceeding the training resolution of diffusion models. The latents is partitioned into smaller regularly cropped regions, which are denoised individually before being reassembled into the final high-resolution output. To reduce visible seams between regions, the cropped regions are overlapped during partitioning. As illustrated in Figure \ref{fig:crop}, the region size and overlap size jointly determine the partitioning strategy and the number of cropped regions. To optimize denoising quality, the region size is typically set to match the training resolution of the diffusion model. The overlap size can be adjusted to balance computational cost and denoising quality. Larger overlap sizes improve denoising quality by reducing boundary artifacts but increase computational cost.
	
	\section{Methodology}
	\subsection{Problem Formulation}
	Given a low-resolution image $I^L$ and corresponding text prompt $p$, the goal is to find an upscaling function $U: I^L \times p \rightarrow {I}^{H}$ such that ${I}^{H}$ satisfies three criteria: (1) \textbf{visual quality}, ensuring ${I}^{H}$ exhibits fewer artifacts and higher fidelity compared to $I^L$; (2) \textbf{global semantic alignment}, where the global structure of ${I}^{H}$ conforms to the semantics implied by $p$ and $I^L$; and (3) \textbf{regional creativity}, where the upscaled image exhibits novel and plausible details at a higher resolution. Here superscripts \textit{L} and \textit{H} denote variables associated with low-resolution and high-resolution, respectively. $I$ represents an image in pixel space. The low-resolution image $I^L$ can be synthesized from text prompt $p$ with a pre-trained diffusion model or captured from the real world. Note that this task differs from typical image super-resolution \cite{dong2023deep,xin2023advanced} or image restoration \cite{zamir2021multi}. Super-resolution aims to remove noise in low-resolution images, such as blur and JPEG artifacts while preserving the structure and details of the low-resolution image. In contrast, our task aims to generate a high-resolution image and encourage the creation of novel and plausible details. Compared to image restoration, where the output resolution is the same as the input, our task requires generating a higher-resolution image while retaining the original structures. Given the semantic and creative nature of the task, we opt to approach this problem by leveraging pre-trained diffusion model $\boldsymbol{\epsilon}_{\boldsymbol{\theta}}(\cdot)$, which is trained on web-scale text-image paired data and contains rich semantics and visual knowledge.
	
	\begin{figure*}[tb]
		\centering
		\includegraphics[width=1.95\columnwidth]{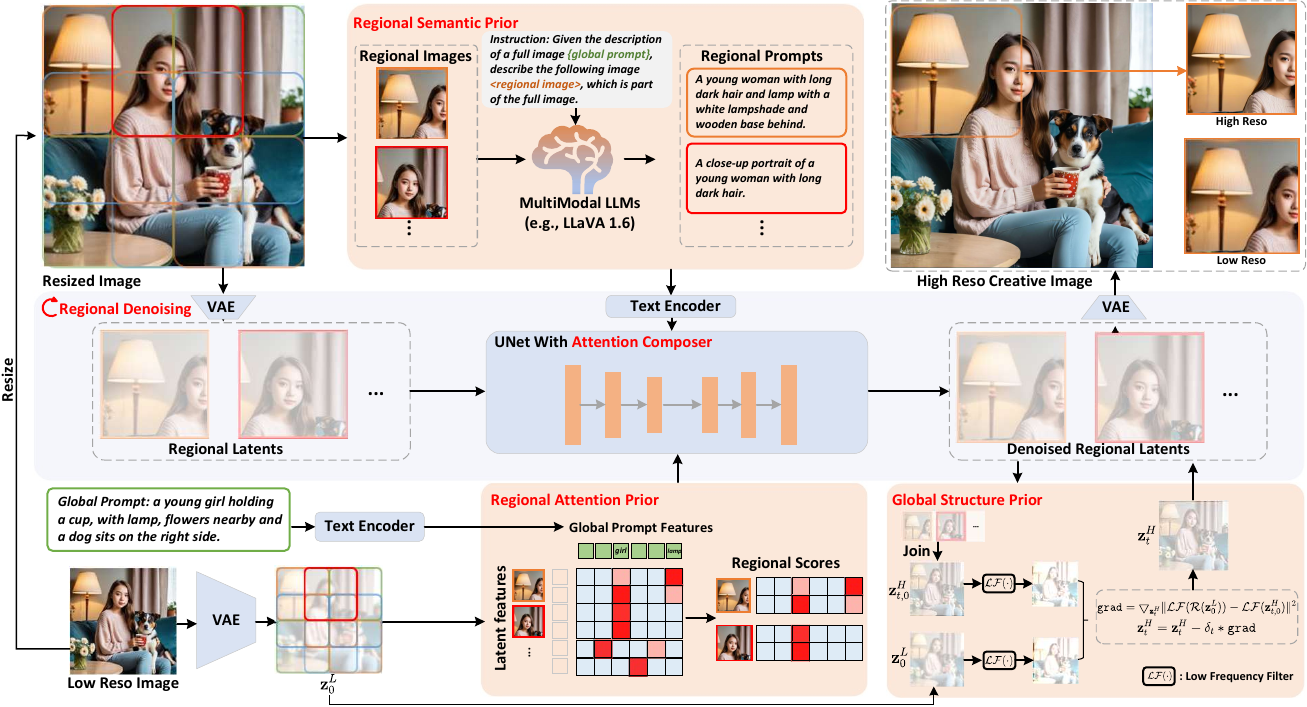}
		\caption{
			The overall framework of C-Upscale for high-resolution generation with global-regional priors. The image is divided into overlapping regions and each region is denoised individually. We leverage Multimodal LLM to generate prompts (Regional Semantic Prior) for each regional image which encourages creative details. Meanwhile, we extract the attention scores (Regional Attention Prior) based on the low-resolution image and global prompt for semantic alignment. These two regional priors are joined with the ``Attention Composer'' module. Finally, the denoised latents at each step are aligned with the low-frequency component of low-resolution latent (Global Structure Prior) to ensure structure alignment. 
		}
		\label{fig:framework}
	\end{figure*}
	
	\subsection{Overall Framework}
	Our overall framework falls into the category of diffuse-then-denoise for upscaling low-resolution images to high resolution. Firstly, the low-resolution image is resized to the target resolution and encoded using a VAE encoder to obtain the latent representation $\mathbf{z}^L$. Next, the diffuse-then-denoise technique is applied to introduce noise to $\mathbf{z}^L$. The noisy latent $\mathbf{z}_t^L$ is then partitioned into cropped regions and each region is denoised individually. These denoised regions are subsequently reassembled to construct the high-resolution latent $\mathbf{z}^H$. Finally, the high-resolution image is reconstructed by decoding $\mathbf{z}^H$ with a VAE decoder. Nevertheless, naively applying regional denoising to cropped regions presents challenges such as misalignment with the global prompt and the absence of global structural guidance. These limitations can lead to object repetition, structure distortions and inaccuracies in detail. To address these issues, we propose \textbf{C-Upscale}, a novel tuning-free image upscaling framework leveraging meticulously designed global-regional priors, as illustrated in Figure 3. The Global Structure Prior (GSP), derived from the low-resolution image, serves as the visual structure's guidance for upscaling. Concurrently, Regional Attention Prior (RAP) is extracted to avoid inconsistencies with the global prompt. To further foster creativity of the diffusion model, we harness Multimodal Large Language Models (MLLMs) to generate Regional Semantic Prior (RSP).
	
	\subsection{Global Structure Prior}
	\label{sec:global}
	As each region undergoes denoising independently, it lacks awareness of the semantic content in other regions, potentially resulting in noticeable seams between regions. Furthermore, the regional latent representations are disrupted by random noise and fail to preserve the original structure of the low-resolution image, making them inadequate for guiding generation. A straightforward approach involves directly utilizing the low-resolution image as a prior and encouraging the high-resolution image to resemble its low-resolution counterpart. However, this may propagate blurriness or graininess defects present in the low-resolution image to the high-resolution result. Acknowledging that the low-frequency component of an image retains its global structure while the high-frequency component encapsulates texture details \cite{graps1995introduction,si2023freeu}, we employ the low-frequency component of $\mathbf{{z}}^L_0$ as the Global Structure Prior (\textbf{GSP}).
	
	Technically, we map the noisy latent $\mathbf{{z}}_t^{H}$ at denoising step $t$ to the clean latent space $\mathbf{{z}}_{t, 0}^H$. We then compute the Mean Squared Error (MSE) between the low-frequency components of the estimated latent and the low-resolution latent $\mathbf{{z}}^L_0$:
	\begin{equation}
		L_{MSE} = \| \mathcal{LF}(\mathcal{R}(\mathbf{{z}}^L_0)) - \mathcal{LF}(\mathbf{{z}}_{t, 0}^H) \|^2.
	\end{equation}
	Here, $R(\cdot)$ performs a resizing operation to match the size of $\mathbf{{z}}_{t, 0}^H$. The operator $\mathcal{LF}(\cdot)$ employs discrete wavelet transformation \cite{graps1995introduction} with the Haar wavelet \cite{stankovic2003haar} to isolate the low-frequency component. To promote alignment between the structure of $\mathbf{{z}}_t^{H}$ and $\mathbf{{z}}^L_0$, we compute the gradient of the $L_{MSE}$ with respect to the latent representation $\mathbf{{z}}_t^{H}$ and update the values of $\mathbf{z}_{t}^{H}$ accordingly.
	\begin{align}
		\texttt{grad} = \bigtriangledown_{\mathbf{z}_{t}^{H}} L_{MSE}, \  \mathbf{z}_{t}^{H} = \mathbf{z}_{t}^{H} - \delta_t * \texttt{grad}.
		\label{eq:gsp}
	\end{align}
	To regulate the strength of the global structure prior, we employ a cosine schedule $\delta_t = s* (1 + \cos[(1-t/T) \pi)]/2$ to gradually diminish the effect of the structure prior as the denoising processes proceeds. Here, $s$ denotes the step size, which also reflects the weight of GSP. In the early stages of denoising, the model prioritizes generating the image's structure, where the low-frequency information plays a pivotal role. As denoising progresses the focus shifts towards generating finer details, the inclusion of the low-frequency prior may disrupt the generation. Thus we lower the influence from the low-frequency component.
	
	\subsection{Regional Attention Prior}
	\label{sec:attention}
	The global structure prior facilitates the alignment of the visual structure in the high-resolution images with that of the original low-resolution image. Nonetheless, a discrepancy persists between the content of the cropped region and the global textual condition, posing the risk of providing misleading guidance for regional denoising. For instance, in Figure \ref{fig:framework}, the red region exclusively depicts a person without a lamp, whereas the global prompt mentions both. Relying on the global prompt to guide the denoising of the red region, will encourage it to generate a lamp in that region erroneously. To resolve this issue, we propose a Regional Attention Prior (\textbf{RAP}) to alleviate the disparity between the cropped region and the global prompt.
	
	SDXL \cite{sdxl} leverages a cross-attention mechanism to incorporate textual conditions into the diffusion process. Specifically, the visual latent representation is flattened into a 1D sequence and linearly projected into a query vector $\mathbf{Q} \in \mathbb{R}^{q \times d}$. Textual feature is also linearly projected into key and value vectors $\mathbf{K} \in \mathbb{R}^{k \times d}$ and $\mathbf{V} \in \mathbb{R}^{k \times d_v}$, respectively. The cross-attention scores are then computed as:
	\begin{equation}
		\mathbf{A} = \text{softmax} (\mathbf{Q} \mathbf{K}^T/ \sqrt{d} ),
	\end{equation}
	where $\mathbf{A}_{ij}$ denotes the relevance between $i^{th}$ spatial latent feature and $j^{th}$ text token. However, for regional denoising, $\mathbf{A}_{ij}$ becomes ambiguous due to inconsistencies between the cropped region and the global prompt. To tackle this issue, we compute the aligned attention score from the low-resolution latent $\mathbf{z}^L_t$ and the global prompt $p$ and extract attention scores for each region. At denoising timestep $t$, we add random noise to low-resolution latent $\mathbf{z}_0^L$ following Eq. \ref{eq:x0_pred} to obtain $\mathbf{z}^L_t$, without requiring access to the generation process of the low-resolution image. Subsequently, we provide it alongside the global prompt $p$ as input to the diffusion model to compute the attention scores $\mathbf{A}_t^L$. To ensure the size of attention scores are consistent with the high-resolution latent, we interpolate the attention map $\mathbf{A}_t^L$ along image dimension and preserve the textual information. Thus we can maintain the distributional coherence and avoid inducing artifacts. Subsequently, for each region, we utilize its coordinates to sample corresponding attention scores serving as regional attention prior. Through the application of regional aligned attention priors, the diffusion model effectively mitigates the risk of attending to incorrect textual conditions, thereby preventing object repetition.
	
	\subsection{Regional Semantic Prior}
	\label{sec:llava}
	
\begin{figure}[h]
		\centering
		\includegraphics[width=\columnwidth]{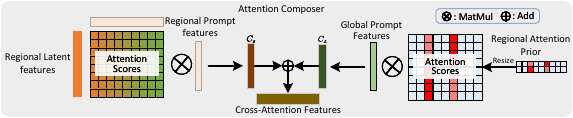}
		\caption{
			Attention Composer between regional semantic prior (left) and regional attention prior (right). On the left, attention computation occurs between regional latent features and regional prompts. On the right, the regional attention scores from the global textual condition are utilized. The attended features from both sides are combined as the final cross-attention features.
		}
		\label{fig:attention}
	\end{figure}

	The global structure prior and regional attention prior serve to align the content of the high-resolution image with the low-resolution image and global prompt $p$. However, they do not explicitly stimulate the creative capacity of the diffusion model. In this section, we delve into further exploiting the potential generation capabilities of diffusion models by directly guiding them to generate detailed regional images. To accomplish this, we leverage MLLMs to generate detailed prompts for each region:
	\begin{equation}
		p_n = MLLMs({I}^{L, (n)}, p), n=1,2, \dots, N.
	\end{equation}
	$I^{L, (n)}$ denotes the $n^{th}$ region of resized low-resolution image and $N$ is the total number of regions. We opt for LLaVA-1.6 \cite{liu2024llavanext} as our MLLMs for its superior performance in generating detailed and diverse descriptions. Note that for regions with limited context (such as sky or road), relying solely on the regional image is insufficient for MLLMs and may lead to ambiguous description. Therefore, in addition to the regional image, the MLLM also takes the global prompt $p$ as input to comprehend the overall image content and avoid generating objects that conflict the overall image. 
	The instruction for LLaVA 1.6 is:
	\begin{tcolorbox}[width=\linewidth, sharp corners=all, colback=white!95!black]
		Instruction: Given the description of a full image {\color{blue}\textbf{\{global prompt\}}}, describe the following image {\color{red}\textbf{\textless regional image\textgreater}} , which is part of the full image.
	\end{tcolorbox}
	\noindent Subsequently, the regional prompts $p_n$ are provided to the diffusion model for each latent patch $\mathbf{z}^{H, (n)}_{t}$, serving as the Regional Semantic Prior  (\textbf{RSP}):
	\begin{equation}
		\mathbf{z}_{t-1}^{H, (n)} = \boldsymbol{\epsilon}_{\boldsymbol{\theta}}(\mathbf{z}^{H, (n)}_t, t, p_n).
	\end{equation}
	\noindent Given that both the regional semantic prior and the regional attention prior influence the cross-attention scores, we have devised an Attention Composer to harness the benefits of both priors. As depicted in Figure \ref{fig:attention}, we initially compute the attention scores between the regional latent and regional prompt to derive the attended semantic features $\mathbf{C}_{s}$. Subsequently, we utilize the attention scores from regional attention prior to obtain the attended features $\mathbf{C}_{a}$. Considering that these two attended features capture slightly different semantic information: the attended feature from the regional attention prior $\mathbf{C}_a$ is closely aligned with the semantic information in the low-resolution image, while the attended feature from the regional semantic prior $\mathbf{C}_s$ emphasizes a more detailed description of the objects within the regional latents. Therefore, to harmonize these two types of semantic information, we average the attended features to yield the final attended features $\mathbf{C}=  (\mathbf{C}_{s} + \mathbf{C}_{a}) / 2$.
	\subsection{Overall Algorithm}
	To provide a clear illustration of how the three priors are combined to facilitate creatively upscaling, we present the overall algorithm of C-Upscale in Algorithm \ref{algo:1}.
	
	\begin{algorithm}[h]
		\caption{C-Upscale}
		\label{algo:1}
		\begin{algorithmic}
			\STATE In this algorithm, we outline the detailed procedure of C-Upscale: $ I^L \times p \rightarrow {I}^{H}$. Here $I^L$ and $p$ denote low-resolution image and corresponding text prompt. $I^H$ represents the upscaled high-resolution image with high quality and fidelity. 
			\STATE \textbf{Require}: Variational autoencoder ($\mathcal{E}$ and $\mathcal{D}$), UNet with Attention Composer $\boldsymbol{\epsilon}_{\boldsymbol{\theta}}(\cdot)$, Multimodal Large Language Models (MLLMs).
			\STATE \textbf{Parameters:} $\overline{\alpha}_i, i=1,\dots, T$ present the predefined schedule in denoising process. $S$ controls the strength of the noise.
			\FOR{$n=1 \text{ to } N$} 
			\STATE \COMMENT{$N$ is total number of regions}
			\STATE ${I}^{L, (n)} = \mathcal{C}(I^L)$ \COMMENT{Crop image into regions	}
			\STATE $ p_n = MLLMs({I}^{L, (n)}, p)$ \COMMENT{RSP}
			\ENDFOR
			\STATE $\mathbf{{z}}^L_0 = \mathcal{E}( I^L)$ \COMMENT{Encode}
			\STATE $\mathbf{{z}}^L_S = \mathcal{N}(\mathbf{{z}}^L_S; \sqrt{\overline{\alpha}_S}\mathbf{{z}}^L_0, ( 1- \overline{\alpha}_S)\mathbf{I}) $ \COMMENT{Add noise}
			\STATE $\mathbf{{z}}^H_S = \mathcal{N}(\mathbf{{z}}^H_S; \sqrt{\overline{\alpha}_S}
			\mathcal{E}(\mathcal{R}(I^L))
			, ( 1- \overline{\alpha}_S)\mathbf{I}) $
			\COMMENT{$\mathcal{R}(\cdot)$ represents resizing}
			\FOR{$t = S \textbf{ to } 1$}
			\STATE $\mathbf{A}_t^L \leftarrow \boldsymbol{\epsilon}_{\boldsymbol{\theta}}(\mathbf{z}^{L}_t, t, p)$ \COMMENT{RAP}
			\FOR{$n =1 \text{ to } N$}
			\STATE $\mathbf{A}_t^{L, (n)} \leftarrow \mathcal{RC}(\mathbf{A}_t^L)$ \COMMENT{Resize and crop}
			\STATE $\mathbf{z}^{H, (n)}_t \leftarrow \mathcal{C}(\mathbf{z}^{H}_t)$ \COMMENT{Crop}
			\STATE  $\mathbf{z}^{H, (n)}_{t-1} = \boldsymbol{\epsilon}_{\boldsymbol{\theta}}(\mathbf{z}^{H, (n)}_t, t, p_n, \mathbf{A}_t^{L, (n)})$ \COMMENT{Attention Composer}
			\ENDFOR
			\STATE $\mathbf{z}^{H}_{t-1} = \mathcal{M} (\mathbf{z}^{H, (n)}_{t-1})$ \COMMENT{Merge separate regions into a complete latent feature}
			\STATE $L_{MSE} = \| \mathcal{LF}(\mathcal{R}(\mathbf{{z}}^L_0)) - \mathcal{LF}(\mathbf{{z}}_{t-1, 0}^H  ) \|^2$ \COMMENT{GSP}
			\STATE $\mathbf{z}_{t-1}^{H} = \mathbf{z}_{t-1}^{H} - \delta_t * \bigtriangledown_{\mathbf{z}_{t-1}^{H}} L_{MSE}$
			\ENDFOR
			\STATE \RETURN $I^H = \mathcal{D}(\mathbf{z}_{0}^{H}) $
		\end{algorithmic}
	\end{algorithm}

	\section{Experiments}
	
	In this section, we evaluate the effectiveness of C-Upscale on randomly sampled test set derived from LAION-5B \cite{schuhmann2022laion} and MS-COCO \cite{lin2014microsoft} for higher-resolution image generation. In Sec. \ref{sec:main results}, we first show both quantitative and qualitative evaluations of C-Upscale compared to other baselines on samples derived from LAION-5B with SDXL \cite{sdxl}. We then examine the generalization ability of C-Upscale on different prompt distribution of MS-COCO and network architecture of SD3 \cite{sd3}, PixArt-$\alpha$ \cite{pixart}, SD-1.5 \cite{ldm} and DreamShaper XL \cite{DreamShaper}. Subsequently, ablation studies are performed in Sec. \ref{sec:emp ana} to validate each design in C-Upscale. We further adapt C-Upscale to more aspect ratios. We also discuss the source of the efficiency gain of C-Upscale compared to DemoFusion and explore efficient ways to implement C-Upscale by leveraging few-step model in Sec. \ref{sec:time}. In Sec. \ref{sec:human} we conduct human preference study to examine C-Upscale against state-of-the-art approaches on three key aspects, i.e., visual quality, global alignment, and regional creativity. Finally, we conduct comparisons between C-Upscale and diffusion-based super-resolution method.

	\subsection{Experimental Settings}
	\noindent \textbf{Setup.}
	In general, we integrate the proposed C-Upscale into different pre-trained diffusion models (i.e., SD 1.5 \cite{ldm}, SDXL \cite{sdxl}, PixArt-$\alpha$ \cite{pixart}, DreamShaper XL \cite{DreamShaper}, Hyper-SDXL \cite{hypersd} and SD3 \cite{sd3}) for higher-resolution image generation. We adopt five image upscaling settings, aiming to upscale the low-resolution image with various upscaling ratios and aspect ratios. During regional denoising, we set the timestep $t$ in the diffuse-then-denoise strategy to be $0.45 * T$, thereby reducing the total number of inference steps by $55\%$ for all models. The actual number of inference steps for each model is set to be $[0.45*N]$, here $N$ denotes the default number of inference steps for each model (SD1.5, SDXL, DreamShaper XL: 50; Hyper-SDXL:8; PixArt-$\alpha$: 20; SD3: 28). Specifically, the actual number of inference steps of Hyper-SDXL is set to $3$. Step size $s$ for gradient updating in GSP is set to $0.2$. The guidance scale is set to $7.5$ for SD1.5, SDXL, Dreamshaper XL and SD3, following DemoFusion \cite{demofusion}. For PixArt-$\alpha$ and Hyper-SDXL, we use the default CFG scales defined in their pipelines, i.e., $4.5$ and $0$. The region size is set to the default size of the training images used by the diffusion models, while the overlap size is set to half of the region size. All experiments were conducted on NVIDIA A100 GPUs.
	\begin{table}[t]
		\resizebox*{\linewidth}{!}{
			\centering
			\begin{tabular}{c|cccccc|c}
				\toprule[1pt]
				Method & MUSIQ $\uparrow$ & ClipIQA $\uparrow$ & ManIQA $\uparrow$ & FID $\downarrow$ & LPIPS $\downarrow$ & CLIP $\uparrow$ & Time \\  \midrule[0.5pt]
				\multicolumn{8}{c}{$2,048 \times 2,048$} \\
				\midrule
				SDXL~\cite{sdxl}      &68.84  & 0.4870     &  0.4432    & 117.11 & - &                   {0.7676} &30sec    \\
				BSRGAN~\cite{zhang2021designing}                 & {71.72} & {0.5291}     & {0.4693}     & \textbf{66.14} & \underline{0.0420} & {0.8584} &1sec  \\
				Real-ESRGAN~\cite{realesr} &\underline{73.41}&0.5210&0.4807&67.25&\textbf{0.0284}& \underline{0.8672} & 3sec \\
				StableSR~\cite{stablesr} &72.26&0.5338&\underline{0.5371}&67.83&0.0558&0.8624&1min\\
				SinSR~\cite{sinsr} &73.02&0.5185&0.5237&68.34&0.0423&0.8583&6sec\\	
				ScaleCrafter~\cite{scalecrafter}             &67.51  & 0.4929     & 0.4540     & 104.57 & -  & 0.7695   & 1min                   \\
				DemoFusion~\cite{demofusion}          &{72.21}  &\underline{0.5346} & {0.5124} &   67.45 & 0.3420  & {0.8525} &2min \\
				C-Upscale                      &\textbf{74.02}  &\textbf{0.5372} & \textbf{0.5534}     &  \underline{66.72} &{0.2766}  & \textbf{0.8687 }      & 1min      \\ \midrule
				\multicolumn{8}{c}{$2,048 \times 4,096$} \\ \midrule
				SDXL~\cite{sdxl}          & 66.34 & 0.4772       & 0.4167      & 161.10 & - &    0.6641 &1min                      \\
				BSRGAN~\cite{zhang2021designing}                      & {66.52}     & 0.4760     & {0.4560}    & 80.11 &  {0.0932}& { 0.7295 }        &    2sec            \\
				Real-ESRGAN~\cite{realesr} &67.12&0.4824&0.4456&\underline{79.84}&\textbf{0.0405}&{0.7900}& 5sec	 \\
				StableSR~\cite{stablesr} &\underline{70.72}&\underline{0.5143}&\underline{0.4703}&80.65&\underline{0.0568}&\textbf{0.8042}&2min\\		
				SinSR~\cite{sinsr} &68.32&0.5075&0.4562&81.63&0.0734&0.7926&15sec\\	
				ScaleCrafter~\cite{scalecrafter}               &  60.77  & {0.5055}     &  {0.4664}     & 228.95 & - &   {0.5654 }   & 5min                   \\
				DemoFusion~\cite{demofusion}           & {69.86} & {0.4797}     & 0.4436     &  {83.18}   &0.3960    & {0.7407  }    &3min                        \\
				C-Upscale                      			 &\textbf{71.34}  &  \textbf{0.5253}    & \textbf{0.4718}     &  \textbf{79.18 }  & {0.3418 }& \underline{0.8008 }           &2min              \\ \midrule
				\multicolumn{8}{c}{$4,096 \times 4,096$} \\
				\midrule
				SDXL~\cite{sdxl}      &64.74  &  0.4367    & 0.4333     &204.09  &-  & 0.6074      &4min                     \\
				BSRGAN~\cite{zhang2021designing}                &{73.46}  & \underline{0.5475}     &{0.5344}      &  \textbf{66.27}& \underline{0.0419} & \underline{0.8701}    & 3sec	                       \\
				Real-ESRGAN~\cite{realesr} &73.39&0.5364&0.4765&67.15&\textbf{0.0349}&0.8597& 8sec \\
				StableSR~\cite{stablesr} &\underline{73.58}&0.5438&\underline{0.5387}&68.02&0.0523&0.8648&4min\\	
				SinSR~\cite{sinsr} &72.64&0.5319&0.5126&68.60&0.0623&0.8593&45sec\\			
				ScaleCrafter~\cite{scalecrafter}             & 59.13 &  0.4657    & 0.4166     &194.91  & -&   {0.4553} &10min \\
				DemoFusion~\cite{demofusion}          &{70.17}  &   {0.4890}   &  0.4757    & 69.64 & 0.3943 &      {0.8677}          & 14min                     \\
				C-Upscale                       			& \textbf{74.23} & \textbf{0.5510}     &  \textbf{0.5594}    & \underline{67.13} &{0.2483}  &    \textbf{0.8726}          &            4min   \\ \midrule[1pt]		
		\end{tabular}}
		\caption{Quantitative comparisons on upscaling synthetic test samples in LAION-5B for higher-resolution image generation. SDXL and ScaleCrafter generate images solely from text prompt without low-resolution inputs, and thus we don't report their LPIPS scores. We mark the best results in \textbf{bold} and the second best results with \underline{underline}.}
		\label{tab:gen}
	\end{table}
	
	\noindent \textbf{Compared Approaches.}
	We compare C-Upscale with several state-of-the-art tuning-free methods and two typical image super-resolution methods: (1) \textbf{Base Model} directly generates higher-resolution images with pre-trained diffusion models without using any upscaling strategy. GAN-based (2) \textbf{BSRGAN} \cite{zhang2021designing}, (3) \textbf{Real-ESRGAN} \cite{realesr} and diffusion-based (4) \textbf{StableSR} \cite{stablesr}, (5) SinSR \cite{sinsr} are widely adopted image super-resolution models that increase image resolution, while strictly retaining content without introducing new details. (6) \textbf{ScaleCrafter} \cite{scalecrafter} generates high-resolution images by exploiting dilated convolution to increase the perceptive field of diffusion models. (7) \textbf{DemoFusion} \cite{demofusion} upgrades regional denoising with skip residual and dilated sampling to encourage global structure alignment. 
	
	\noindent \textbf{Dataset.}
	Since our C-Upscale is formulated as a unified image upscaler, it is feasible to upscale both synthetic and real-world low-resolution images. To fully validate the robustness of our C-Upscale, we leverage two different kinds of inputs (i.e., synthetic and real-world images) for evaluation. Specifically, for \textbf{synthetic} test images, we first randomly sample 1K prompts from LAION-5B \cite{schuhmann2022laion} dataset and use pre-trained diffusion models to generate a set of low-resolution images as inputs for upscaling. Another 1K prompts are also randomly sampled from MS-COCO \cite{lin2014microsoft} to generate a set of low-resolution images. For \textbf{real-world} test images, we directly use the ground-truth images corresponding to previously sampled 1K LAION-5B prompts, and resize-and-crop them into deisred lower-resolution images.
	
	\begin{table}[t]
		\resizebox*{\linewidth}{!}{
			\centering
			\begin{tabular}{c|cccccc}
				\toprule[1pt]
				Method & MUSIQ $\uparrow$ & ClipIQA $\uparrow$ & ManIQA $\uparrow$ & FID $\downarrow$ & LPIPS $\downarrow$ & CLIP $\uparrow$ \\  \midrule[0.5pt]
				\multicolumn{7}{c}{$2,048 \times 2,048$} \\  
				\midrule
				BSRGAN~\cite{zhang2021designing}                 &62.59  & 0.4023     & \underline{0.4374}     &    \textbf{39.00}     & \underline{0.1343} & 0.8862      \\
				Real-ESRGAN~\cite{realesr} &62.94&0.4261&0.4168&\underline{41.05}&\textbf{0.0948}&0.8769  \\
				ScaleCrafter~\cite{scalecrafter}             & 57.19 & 0.3467     & 0.2798     &   48.28     &0.2973  & \underline{0.8887} \\
				DemoFusion~\cite{demofusion}          &\underline{66.43} & \underline{0.4499} & 0.4145  &46.15 	&{0.3154	} & {0.8862}             \\
				C-Upscale                      &\textbf{69.87} & \textbf{0.4668} & \textbf{0.4661}    &{42.22} &{0.2706}  &\textbf{0.8940}        \\ \midrule
				\multicolumn{7}{c}{$2,048 \times 4,096$} \\ \midrule
				BSRGAN~\cite{zhang2021designing}              & 55.26 & 0.3419     &  \underline{0.3473}    & {54.09} & \underline{0.1916} & \underline{0.7979}                          \\
				Real-ESRGAN~\cite{realesr} &56.42&0.3514&0.3356&53.37&\textbf{0.1573}& 0.7954 \\
				ScaleCrafter~\cite{scalecrafter}             & 44.14     & 0.3058     & 0.2378  &  56.88&  0.3296 &{0.7939}         \\
				DemoFusion~\cite{demofusion}    & \underline{57.77} & \underline{0.4272}     &  0.3214&\underline{48.92} &  0.3261  & {0.7974}         \\
				C-Upscale                       & \textbf{63.95}&  \textbf{0.4847}    & \textbf{0.4021}     &\textbf{48.59}  & {0.3107} &       \textbf{0.8003}                   \\ \midrule
				
				\multicolumn{7}{c}{$4,096 \times 4,096$} \\
				\midrule
				BSRGAN~\cite{zhang2021designing}              & 62.53 & 0.4513     & 0.4380 & \underline{41.98}     &  \textbf{0.1429}     & \textbf{0.8867}                        \\
				Real-ESRGAN~\cite{realesr} &63.55&0.4671&0.4394&42.08&\underline{0.1596}&0.8854  \\
				ScaleCrafter~\cite{scalecrafter}             & 52.47 & 0.3728     & 0.2546     &   68.10    & 0.4936 &  {0.8110}      \\
				DemoFusion~\cite{demofusion}          & \underline{67.15} &  \underline{0.4622}    &   \underline{0.4410 }  &   51.24    & 0.3521 &    0.8672                                 \\
				C-Upscale                       &\textbf{69.70}  &\textbf{0.4765 }       &  \textbf{0.4607}    &   \textbf{41.42}    & {0.2522} &  \underline{0.8862}                       \\ \midrule[1pt]				                    
		\end{tabular}}
		\caption{Quantitative comparisons on upscaling real-world test samples in LAION-5B for higher-resolution image generation. We mark the best results in \textbf{bold} and the second best results with \underline{underline}.}
		\label{tab:real}	
	\end{table}
	
	\subsection{Main Results}
	\label{sec:main results}
	\noindent \textbf{Quantitative Evaluation.}
	Here we show the quantitative evaluation of our C-Upscale in higher-resolution image generation under three settings with varied aspect ratios (i.e., 2,048$\times$2,048, 2,048$\times$4,096 and 4,096$\times$4,096). Several representative metrics are adopted: non-reference image quality metrics (MUSIQ \cite{ke2021musiq}, ClipIQA \cite{wang2023exploring} and ManIQA \cite{yang2022maniqa}) and full-reference image quality metrics (FID \cite{heusel2017gans} and LPIPS \cite{zhang2018perceptual}), and semantic relevance metric (CLIP \cite{radford2021learning}). Concretely, the first three non-reference image quality metrics are leveraged to evaluate the perceptual quality of images according to human preference. They focus on multiple key aspects like perceptual distortions, sharpness, naturalness, and artistic value. FID measures the distance between synthetic and ground-truth data distributions. LPIPS assesses the perceptual similarity between input low-resolution images and output higher-resolution images. CLIP score reflects the global semantic relevance between global prompts and generated images. Notably, all runs here capitalize on the same diffusion model (SDXL) for fair comparison.	
	
	Table \ref{tab:gen} summarizes the performance comparisons for upscaling sampled synthetic samples in LAION-5B. Note that both SDXL and ScaleCrafter generate higher-resolution images solely from text prompt, without using low-resolution synthetic images. Thus their pairwise perceptual similarity scores (i.e., LPIPS) are not reported. Overall, our C-Upscale consistently achieves better performances against existing tuning-free techniques (SDXL, ScaleCrafter, DemoFusion) and image super-resolution methods (BSRGAN, Real-ESRGAN, StableSR and SinSR) across most metrics for each upscaling setting. Remarkably, under ultra-high resolution (4,096$\times$4,096), C-Upscale achieves absolute non-reference and full-reference image quality scores (MUSIQ and FID) improvement of 4.06\% on MUSIQ and 2.51\% on FID against the best tuning-free diffusion-based model (DemoFusion). Such clear performance boosts demonstrate the key advantage of global-regional priors for high-quality and creative generation of high-resolution images. 
	
	Specifically, SDXL enables tuning-free generation of higher-resolution images via direct sampling. But the training-inference discrepancy on image resolution commonly results in degraded higher-resolution image generation with severe object repetition and distortion issues, resulting in degraded image quality scores. 
	BSRGAN, Real-ESRGAN, StableSR and SinSR regard image upscaling as typical super-resolution task that strictly retains visual content of low-resolution images and thus eliminates object repetition and distortion issues with better image quality. Therefore they achieve better performances than recent diffusion-based tuning-free approaches (ScaleCrafter and DemoFusion) in most image quality metrics, where the degenerated results of the tuning-free approaches might be caused by the stochasticity in diffusion model which somewhat hurts the preservation of visual appearance. It is noteworthy that, C-Upscale, by guiding the creative image upscaling with global structure priors, alleviates the downsides and obtains comparable full-reference image quality metrics against the super-resolution methods. However, the typical super-resolution task tends to introduce regional smoothing especially when the resolution scales up, instead of adding creative regional details, potentially leading to slightly inferior image quality than C-Upscale. \\
	\indent Considering the use of pre-trained diffusion backbone in C-Upscale, we also examine the performance of the diffusion-based super-resolution models (i.e., StableSR and SinSR). Results show that C-Upscale again exhibits better performances in image quality metrics. We speculate that the phenomenon might be derived from the conventional training paradigm of the super-resolution models. Super-resolution models are all trained on synthetic datasets, where low-resolution images are generated by applying degradation processes like downsampling and JPEG compression to high-resolution images. As a result, the creative nature of the diffusion models is significantly constrained after training under super-resolution paradigm. That is, the image details generated by diffusion-based super-resolution methods are inherently shaped and constrained by the specific degradation methods to remove blurriness and JPEG artifacts. In contrast, our C-Upscale leverages diffusion models (pre-trained on web-scale datasets) in a tuning-free manner, offering greater flexibility in generating creative details.
    
	It is noteworthy that in the 2048$\times$4096 setting, C-Upscale outperforms super-resolution methods on FID metric. This is because the synthetic rectangular low-resolution images are of poorer quality than the square images due to the SDXL training-inference discrepancy, and super-resolution methods adhere too closely to these artifacts due to its regional smoothing property. In contrast, our C-Upscale effectively recovers these artifacts through regional denoising with rich details, thereby achieving superior results. When comparing to ScaleCrafter and DemoFusion, our C-Upscale facilitates the generation of high-fidelity and creative regional details with additional regional attention prior and regional semantic prior, thereby achieving better non-reference image quality scores.
	
	\begin{figure*}[h]
		\centering
		\includegraphics[width=1.99\columnwidth]{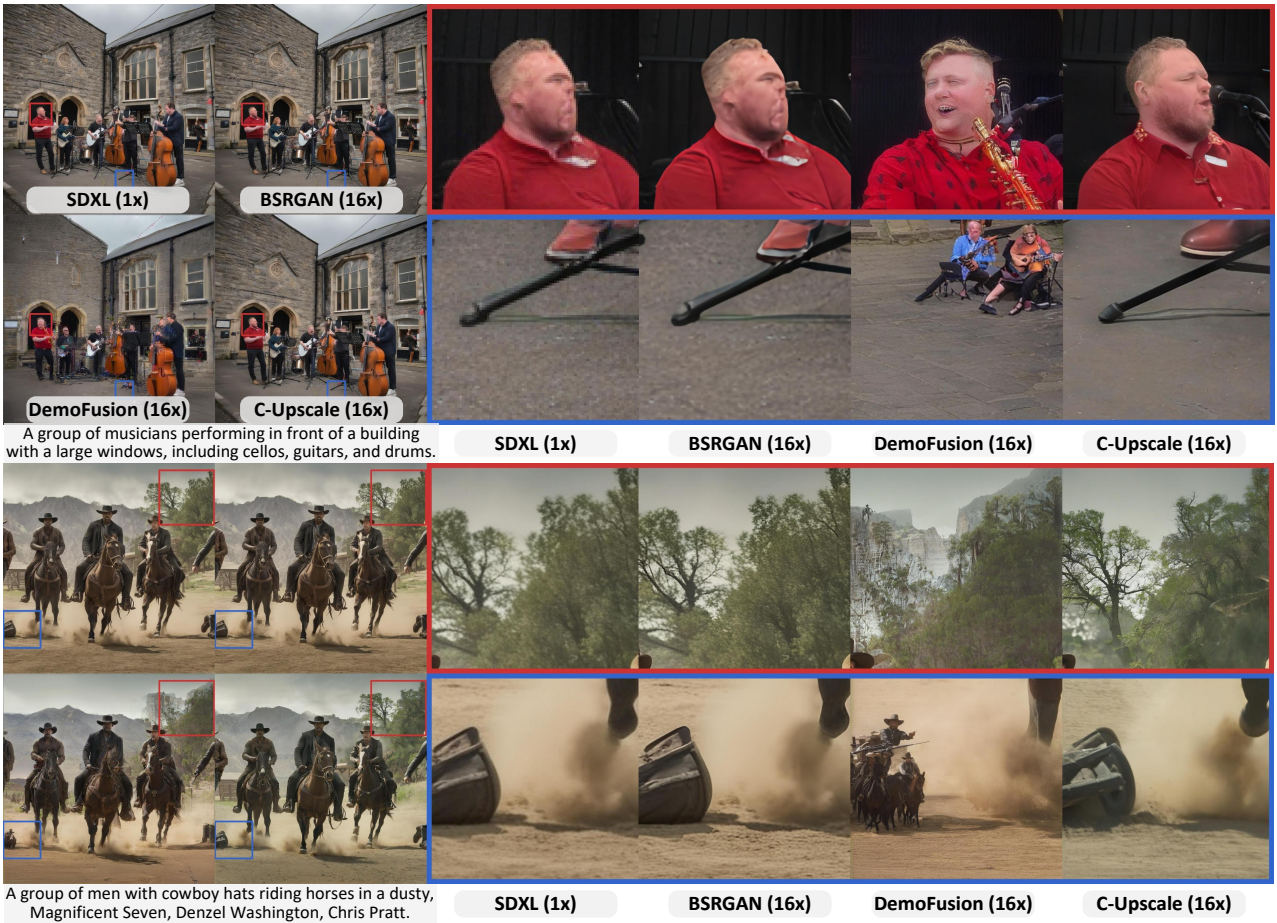}
		\caption{Two visual examples of upscaling the low-resolution images (1,024$\times$1,024 outputs of SDXL) into higher-resolution ones (4,096$\times$4,096) by different tuning-free methods. The regions in red and blue boxes are presented in zoom-in view.}
		\label{fig:compare}
		\vspace{-0.5cm}
	\end{figure*}

	Table \ref{tab:real} details the performance comparisons when upscaling real-world test samples in LAION-5B. It is worth noting that ScaleCrafter is primarily designed for text-to-image generation, without the requirement of input low-resolution inputs. Thus we remould ScaleCrafter via ``diffuse-then-denoise'' paradigm to permit the real-world low-resolution inputs. In particular, we first convert the input low-resolution image into an intermediate state by adding noise in the diffusion process, and subsequently denoise it by using ScaleCrafter to obtain the final output image. Note that we adopt the same noise strength (0.45) as in our C-Upscale for fair comparison. Similar to the observations on upscaling synthetic test samples, our C-Upscale also exhibits better performances than other approaches for most non-reference and full-reference image quality metrics. The results again demonstrate the merit of global-regional priors for higher-resolution image generation.

	\noindent \textbf{Qualitative Comparisons.} Figure \ref{fig:compare} showcases image upscaling results from 1,024$\times$1,024 SDXL outputs to higher-resolution outputs (4,096$\times$4,096) for two examples. It is easy to see that all three upscalers can generate higher-resolution images with somewhat enhanced quality, while C-Upscale produces higher-fidelity images with more creative regional details by jointly exploiting global-regional priors. Taking the upper results as an example, the image super-resolution method (BSRGAN) introduces regional smoothing without adding new details (see the over-smoothed face in red box), and DemoFusion hallucinates completely different people in red box. Instead, our C-Upscale produces visually similar people with clearer eye and nose contours. Moreover, DemoFusion suffers from object repetition issue (see the miniature people playing instruments in blue box), while C-Upscale eliminates this issue with better global structure alignment.
	
	\begin{figure*}[t]
		\centering
		\includegraphics[width=2\columnwidth]{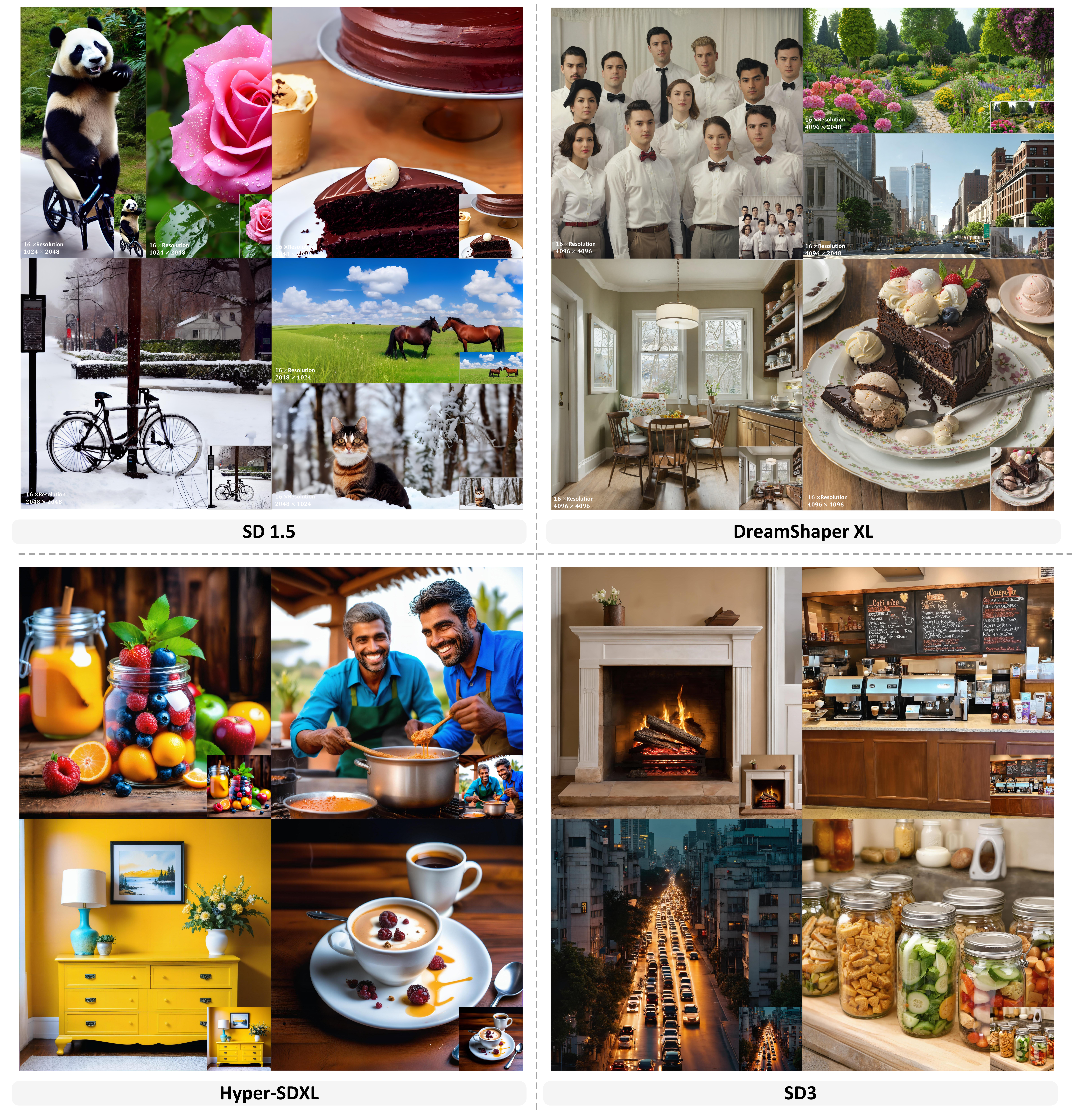}
		\caption{Visual examples of our C-Upscale in four different diffusion models (i.e., SD 1.5, DreamShaper XL, Hyper-SDXL and SD3) for 16$\times$ image upscaling. Note that the images denoted by Hyper-SDXL are generated using our accelerated C-Upscale integrated with Hyper-SDXL and vLLM.}
		\label{fig:sd}
		\vspace{-0.5cm}
	\end{figure*}
	
	\begin{table}[t]
		\resizebox*{\linewidth}{!}{
			\centering
			\begin{tabular}{c|cccccc}
				\toprule[1pt]
				Method & MUSIQ $\uparrow$ & ClipIQA $\uparrow$ & ManIQA $\uparrow$ & FID $\downarrow$ & LPIPS $\downarrow$ & CLIP $\uparrow$ \\  \midrule[0.5pt]
				\multicolumn{7}{c}{$2,048 \times 2,048$} \\  
				\midrule
				SDXL~\cite{sdxl} &70.15&0.5208&0.4314&99.54&-& 0.8049 \\ 
				BSRGAN~\cite{zhang2021designing}       &71.22  & 0.5300  &0.4803&\textbf{65.45}&\textbf{0.0074}&\underline{0.8340}    \\
				Real-ESRGAN~\cite{realesr} &71.54&0.5348&0.4852&66.48&\underline{0.0241}& 0.8338 \\
				ScaleCrafter~\cite{scalecrafter}    &\underline{72.56}&0.5397&0.5250&86.47&-&0.7661  \\
				DemoFusion~\cite{demofusion}      &{72.40} & \underline{0.5479} &\underline{0.4891}&67.84&0.2469&  0.8315 \\
				C-Upscale     &\textbf{74.53} & \textbf{0.5656}&\textbf{0.5406}&\underline{65.98}&{0.1662}&  \textbf{0.8350}      \\ 
				\midrule
				\multicolumn{7}{c}{$2,048 \times 4,096$} \\ \midrule
				SDXL~\cite{sdxl} &67.06&0.4829&0.4149&141.24&-&0.6372  \\
				BSRGAN~\cite{zhang2021designing}       &70.48&0.5148&0.4358&80.73&\textbf{0.0791}&0.7824  \\
				Real-ESRGAN~\cite{realesr} &\underline{70.94}&\underline{0.5196}&\underline{0.4385}&\underline{79.56}&\underline{0.0974}&\underline{0.7854}  \\
				ScaleCrafter~\cite{scalecrafter}      &69.84&0.5049&0.4152&162.45&-&0.7428  \\
				DemoFusion~\cite{demofusion}  &70.30&0.5131&0.4240&81.54&0.2756&0.7728  \\
				C-Upscale                     &\textbf{71.80}&\textbf{0.5271}&\textbf{0.4500}&\textbf{78.42}&0.2050&\textbf{0.7979}  \\ 
				\midrule		
				\multicolumn{7}{c}{$4,096 \times 4,096$} \\
				\midrule                 
				SDXL~\cite{sdxl} 						&67.52&0.4176&0.4053&147.56&-&0.6260  \\ 
				BSRGAN~\cite{zhang2021designing}       	&\underline{72.51}&0.5547&\underline{0.5288}&\underline{65.48}&\textbf{0.0077}&\underline{0.8359}  \\
				Real-ESRGAN~\cite{realesr} 				&72.17&0.5319&0.5248&66.84&\underline{0.0233}&0.8345  \\
				ScaleCrafter~\cite{scalecrafter}  		&70.21&0.4894&0.4574&154.17&0.4512&0.8244  \\
				DemoFusion~\cite{demofusion}     		&71.35&0.5082&0.4497&67.54&0.3937&0.8259  \\
				C-Upscale   							&\textbf{74.52}&\textbf{0.5717}&\textbf{0.5486}&\textbf{65.24}&0.2055&\textbf{0.8408 } \\ \midrule[1pt]				                    
		\end{tabular}}
		\caption{Quantitative comparisons on upscaling synthetic test samples in MS-COCO for higher-resolution image generation. SDXL and ScaleCrafter generate images solely from text prompt without low-resolution inputs, and thus we don't report their LPIPS scores. We mark the best results in \textbf{bold} and the second best results with \underline{underline}.}
		\label{tab:coco}
		\vspace{-0.5cm}
	\end{table}
	
	\begin{table}[t]
		\resizebox*{\linewidth}{!}{
			\renewcommand{\arraystretch}{1.17}
			\centering
			\begin{tabular}{c|cccccc}
				\toprule[1pt]
				Method & MUSIQ $\uparrow$ & ClipIQA $\uparrow$ & ManIQA $\uparrow$ & FID $\downarrow$ & LPIPS $\downarrow$ & CLIP $\uparrow$ \\  \midrule[0.5pt]
				\multicolumn{7}{c}{$2,048 \times 2,048$} \\  
				\midrule
				PixArt-$\alpha$~\cite{pixart} &59.10&0.4841&0.3709&86.74&-&0.6738  \\ 
				BSRGAN~\cite{zhang2021designing}     &73.58&0.6045&0.5789&68.54&0.0842&0.7684  \\ 
				Real-ESRGAN~\cite{realesr} &\underline{73.64}&0.6184&0.5768&\underline{67.95}&0.0549&\underline{0.7759}  \\
				DemoFusion~\cite{demofusion}  &72.59&\underline{0.6288}&\underline{0.5812}&69.87&0.3806&0.7725  \\ 
				C-Upscale &\textbf{74.27}&\textbf{0.6478}&\textbf{0.6049}&\textbf{67.84}&0.1932&\textbf{0.7861}  \\ 
				\midrule
				\multicolumn{7}{c}{$2,048 \times 4,096$} \\ \midrule
				PixArt-$\alpha$~\cite{pixart} 			&53.15&0.4050&0.4240&142.14&-&0.7235  \\
				BSRGAN~\cite{zhang2021designing}       	&72.45&\underline{0.6057}&\underline{0.5364}&67.54&\underline{0.0972}&\underline{0.7563}  \\
				Real-ESRGAN~\cite{realesr}	 			&\underline{72.95}&0.5984&0.5266&\underline{67.12}&\textbf{0.0598}&0.7529  \\
				DemoFusion~\cite{demofusion}  			&68.01&0.5997&0.4816&71.54&0.3589&0.7329  \\
				C-Upscale                     			&\textbf{73.23}&\textbf{0.6347}&\textbf{0.5544}&\textbf{66.51}&0.1772&\textbf{0.7573}  \\ 
				\midrule		
				\multicolumn{7}{c}{$4,096 \times 4,096$} \\
				\midrule                 
				PixArt-$\alpha$~\cite{pixart} 		&51.32&0.4606&0.4534&121.25&-&0.5606  \\ 
				BSRGAN~\cite{zhang2021designing}    &72.55&\underline{0.6017}&0.5849&\underline{66.85}&\textbf{0.0511}& \underline{0.7698} \\
				Real-ESRGAN~\cite{realesr}	 		&\underline{73.24}&0.5984&\underline{0.5974}&67.84&\underline{0.1052}&0.7586  \\
				DemoFusion~\cite{demofusion}     	&69.78&0.5808&0.5460&70.54&0.5209&0.7310  \\
				C-Upscale   						&\textbf{73.88}&\textbf{0.6284}&\textbf{0.6133}&\textbf{65.57}&{0.1668}&\textbf{0.7856}  \\ \midrule[1pt]				                    
		\end{tabular}}
		\caption{Quantitative comparisons on upscaling synthetic test samples in LAION-5B for higher-resolution image generation. Here PixArt-$\alpha$ generate images solely from text prompt without low-resolution inputs, and thus we don't report their LPIPS scores. We mark the best results in \textbf{bold} and the second best results with \underline{underline}.}
		\label{tab:pixart}
		\vspace{-0.4cm}
	\end{table}
	
	\noindent \textbf{Generalization of C-Upscale.} 
	Here, we study the generalization ability of the proposed C-Upscale across different prompt distributions and network architectures. In addition to the prompts from LAION-5B, we report quantitative results based on sampled captions from the MS-COCO dataset. Furthermore, we adapt C-Upscale to the recently proposed Transformer-based diffusion model, PixArt-$\alpha$ \cite{pixart} and SD3 \cite{sd3}, as well as to UNet-based models, i.e., SD 1.5 \cite{ldm} and DreamShaper XL \cite{DreamShaper}.
	
	Table \ref{tab:coco} presents performance comparisons for upscaling synthetic images generated with MS-COCO prompts. Compared to LAION-5B prompts, MS-COCO prompts are generally more concise and formal. The performance trends observed are similar to those with LAION-5B prompts, validating the generalization of C-Upscale across different prompt distributions.
	
	For the experimental results on PixArt-$\alpha$, we first generated low-resolution images using the pre-trained PixArt-$\alpha$-1024 checkpoint with LAION-5B prompts. Since our C-Upscale is a general tuning-free framework for creative upscaling and independent of specific diffusion architectures, we can effortlessly adapt it to the PixArt-$\alpha$ framework and upscale input images to various target resolutions. For fair comparisons with other approaches, we also integrated tuning-free method DemoFusion into PixArt-$\alpha$ and optimized its hyper-parameters to the best of our ability. We exclude Scalecrafter for comparison as it is specifically designed for convolutional architectures. Table \ref{tab:pixart} summarizes the performance comparisons. Similar to observations on the SDXL base model, our C-Upscale exhibits better results than other approaches on most metrics, demonstrating its generalization to the latest diffusion architectures. 
	
	We also integrate our C-Upscale into three additional diffusion models, SD 1.5 \cite{ldm}, DreamShaper XL \cite{DreamShaper} and SD3 \cite{sd3}. Figure \ref{fig:sd} shows 16$\times$ upscaling results using our C-Upscale in these diffusion models. Consistent enhancements in visual fidelity and regional creativity are observed in each diffusion model, further validating the effectiveness of C-Upscale.
	
	\begin{figure*}[h]
		\centering
		\includegraphics[width=2\columnwidth]{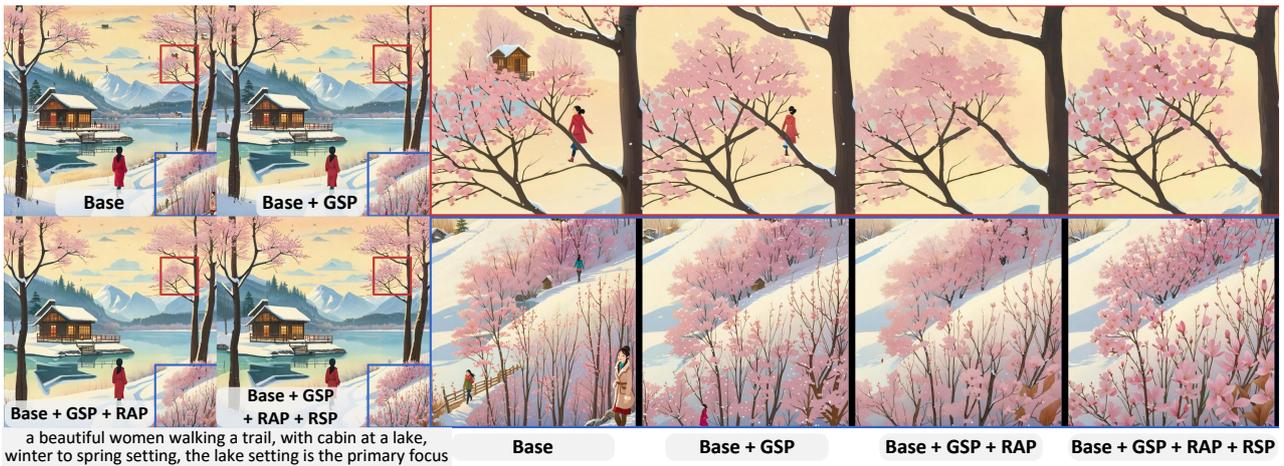}
		\caption{Ablation study by considering different prior for 16$\times$ image upscaling. The regions in red and blue boxes are presented in zoom-in view for comparison.}
		\vspace{-0.3cm}
		\label{fig:ablation}
	\end{figure*}

	\begin{table}
			\resizebox*{\linewidth}{!}{
				\centering
				\begin{tabular}{c|cccccc}
					\toprule[1pt]
					Method & MUSIQ $\uparrow$ & ClipIQA $\uparrow$ & ManIQA $\uparrow$ & FID $\downarrow$ & LPIPS $\downarrow$ & CLIP $\uparrow$ \\  \midrule[0.5pt]
					-GSP &\underline{73.11}&\underline{0.5222}&0.5464&\underline{66.81}&0.2793&\underline{0.8677}  \\ 
					-RAP   &73.10&0.5187&\underline{0.5471}&67.12&{0.2794}&0.8608  \\ 
					-RSP &72.80&0.5200&0.5426&67.62&\underline{0.2782}&0.8672  \\
					\midrule
					C-Upscale &\textbf{74.02}  &\textbf{0.5372} & \textbf{0.5534}     &  \textbf{66.72} &\textbf{0.2766}  & \textbf{0.8687 }  \\ 
					\midrule[1pt]			                    
			\end{tabular}}
			\caption{Ablation studies on removing the three priors from C-Upscale respectively. All images are upscaled from synthetic test samples in LAION-5B to 2,048$\times$2,048.}
			\label{tab:ablation} \vspace{-0.5cm}
		\end{table}
		
		\subsection{Experimental Analysis}
		\vspace{-0.3cm}
		\label{sec:emp ana}
		\noindent \textbf{Ablation Study on Three Priors.}
		Here we study how each prior in our C-Upscale influences the image quality of upscaled higher-resolution image both quantitatively and qualitatively. We remove each prior from the complete C-Upscale respectively and denote each setting as ``-X'', where ``X'' indicates the removed prior. The comparisons presented in Table \ref{tab:ablation} indicate that removing Global Structure Prior reduces the perceptual similarity between the input low-resolution images and the high-resolution images. This validates the efficacy of GSP in preserving content and structure of low-resolution images. The drop in performance induced by removing Regional Attention Prior (RAP) confirms the RAP's effectiveness in aligning the high-resolution images to the given prompt. The degraded image quality scores (MUSIQ and ManIQA) further demonstrate that Regional Semantic Prior (RSP) can indeed enhance image quality by stimulating creative details.We also assess the contribution of each prior qualitatively. As illustrated in Figure \ref{fig:ablation}, the incremental addition of priors reduces regional repetitions and enhances creative details. These quantitative and qualitative results collectively demonstrate the complementary nature of various priors.

\begin{figure}[tb]
			\centering
			\includegraphics[width=.95\columnwidth]{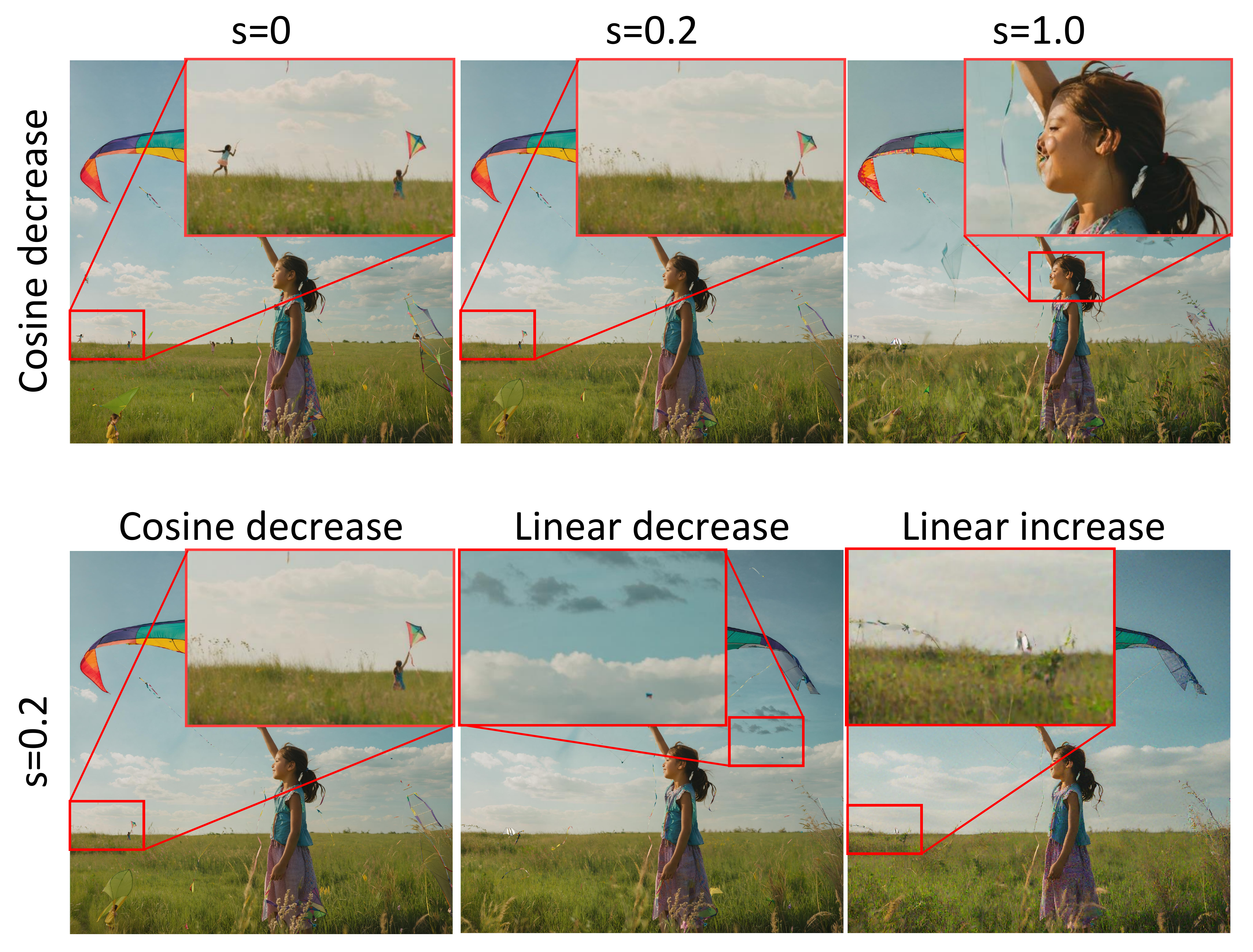}
			\caption{
				Image upscaling with different weights and schedulers in Global Structure Prior.
			}
			\label{fig:gsp}
		\end{figure}

		\noindent \textbf{Ablation Study on Global Structure Prior (GSP).} 
		We proceed to examine the effect of different step sizes and schedules (defined in Eq. \ref{eq:gsp}) on preserving image structure from the low-resolution image. Specifically, we disable regional attention prior and regional semantic prior, and solely evaluate the impact of GSP with three different step sizes/schedulers. First, we set the scheduler to a decreasing cosine function and vary the step size $s$ (also can be regarded as the weight of GSP) as 0, 0.2, and 1.0. As shown in the first row of Figure \ref{fig:gsp}, increasing $s$ effectively reduces repeated objects, highlighting the role of GSP in guiding structure synthesis. Nevertheless, an excessively large step size $s = 1.0$ introduces visible artifacts on the human face, as the overemphasis on GSP disrupts original model's predictions. Next, we evaluate two alternative schedulers with a fixed step size of 0.2: a linearly decreasing scheduler (\( \delta_t = s \cdot t / T \)) and an increasing scheduler (\( \delta_t = s \cdot (1 - t / T) \)). Note that the timestep $t$ flows from $T$ to $0$. Linear decreasing scheduler exhibits fewer repetitions but produce artifacts (see the black clouds which are absent in the original image). Conversely, linear increasing scheduler incurs noticeable blurriness. We choose the cosine scheduler with 0.2 weight in our experiments. While the cosine scheduler exhibits some repetitions, these can be mitigated with the regional attention prior. 
		
\begin{figure}[h]
			\vspace{-0.3cm}
			\centering
			\includegraphics[width=1.0\columnwidth]{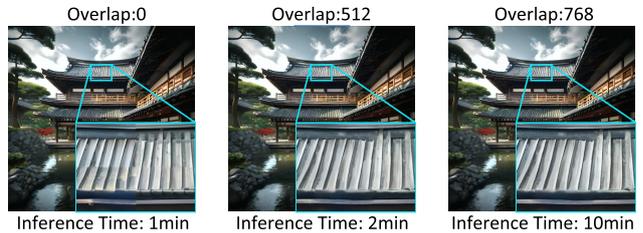}
			\caption{
				Various overlap sizes and corresponding inference time. All images are upscaled to 4,096$\times$4,096 using SDXL.
			}
			\label{fig:stride}\vspace{-0.4cm}
		\end{figure}

		\noindent \textbf{Region Size and Stride.}
		The region size in Figure \ref{fig:framework} is set as default resolution of the base diffusion models (e.g. 1,024 of SDXL) to fully utilize base model's potential. Following \cite{multidiffusion}, we incorporate overlaps between regions to mitigate seams among them. Larger overlaps reduce visible artifacts at the boundaries but increase inference time. We vary overlap size within $\{0,512,768\}$ in Figure \ref{fig:stride} to demonstrate this phenomenon. No discrepancies are observed when the overlap size exceeds 512, but inference time increases drastically when the overlap size reaches 768. Therefore, we opt for an overlap size of 512 for SDXL throughout this paper.	

		\noindent \textbf{More Aspect Ratios.}
		To further validate the robustness of C-Upscale on irregular image sizes, we implement C-Upscale on two more aspect ratios, i.e., 1,536$\times$2,048 and 4,096$\times$2,048. Table \ref{tab:aspect-ratio} summarizes the results. SDXL tends to produce low-resolution images with poorer quality on these aspect ratios (3:4 and 2:1) due to their lower representation in the training distribution. Despite this, C-Upscale exhibits superior performance compared to super-resolution methods and other diffusion-based upscaling methods, proving its efficacy on low-quality input images.
		
		\begin{table}[h]
			\resizebox*{\linewidth}{!}{
				\begin{tabular}{c|cccccc}
					\toprule[1pt]
					Method & MUSIQ $\uparrow$ & ClipIQA $\uparrow$ & ManIQA $\uparrow$ & FID $\downarrow$ & LPIPS $\downarrow$ & CLIP $\uparrow$ \\  \midrule[0.5pt]
					\multicolumn{7}{c}{$1,536 \times 2,048$} \\  
					\midrule\
					SDXL~\cite{sdxl} &71.55&0.5551&0.4878&79.54&-&0.7979\\
					BSRGAN \cite{zhang2021designing}  & \underline{74.83}& \underline{0.5705}& \underline{0.5481}&\underline{65.51}&\textbf{0.0173}&0.8472   \\
					Real-ESRGAN~\cite{realesr} &74.29&0.5674&0.5387&66.16&\underline{0.0524}&0.8468  \\ 
					DemoFusion  \cite{demofusion}       &{72.76}&0.5613&0.5153&66.28&0.2703& \underline{0.8486}              \\
					C-Upscale          &\textbf{75.22}&\textbf{0.5857}&\textbf{0.5725}&\textbf{65.06}&{0.1776}&\textbf{0.8501}   \\ \midrule
					\multicolumn{7}{c}{$4,096 \times 2,048$} \\ \midrule
					SDXL~\cite{sdxl}  &67.86&0.5301&0.4612&105.48&-&06436  \\ 
					BSRGAN~\cite{zhang2021designing}            & 72.52 &0.5178&0.4658&91.44&\textbf{0.0303}&   0.7690        \\
					Real-ESRGAN~\cite{realesr} &\underline{72.78}&0.5247&0.4585&91.27&\underline{0.0598}&0.7654  \\ 
					DemoFusion~\cite{demofusion}   & {71.34} &\underline{0.5443}&\underline{0.4728}&\underline{90.77}&0.3047&\underline{0.7700}   \\
					C-Upscale                       & \textbf{73.50}&\textbf{0.5633}&\textbf{0.5205}&\textbf{89.78}&{0.2115}&\textbf{0.7725}  \\
					\midrule[1pt]				                    
			\end{tabular}}
			\caption{Experiments on two more aspect ratios. The low-resolution input images are synthetic samples generated by SDXL from with LAION-5B prompts.}
			\label{tab:aspect-ratio}\vspace{-0.5cm}
		\end{table}
		
		\subsection{Computational Efficiency} \label{sec:time}
		\textbf{Inference Time Analysis.} Apart from generating higher-resolution images with enhanced global semantic alignment and regional creativity, our C-Upscale demonstrates superior computational efficiency compared to other diffusion-based image upscaling approaches. Compared to DemoFusion, the inference time advantage of C-Upscale stems from two points: the cost-expensive design of \textit{progressive upscaling} in DemoFusion, and the cost-efficient \textit{diffuse-then-denoise} with less steps in C-Upscale. For instance, when upscaling an image from 1,024$\times$1,024 to 4,096$\times$4,096, DemoFusion employs a progressive strategy, sequentially upscaling to 2,048$\times$2,048, 3,072$\times$3,072, and finally to the target resolution, which is inherently slower than directly upscaling from 1,024$\times$1,024 to 4,096$\times$4,096 in C-Upscale. Furthermore, DemoFusion generates high-resolution images from pure Gaussian noise, requiring 50 steps for SDXL. In contrast, C-Upscale utilizes the diffuse-then-denoise technique, reducing the number of steps to 22 steps.
		
		We further give a more detailed calculation on the computational cost regarding each modules. Overall, the most time-consuming operations of diffusion-based upscaling are UNet denoising and prompt generation with LLaVA in our C-Upscale. The computational cost of UNet denoising scales linearly with both the number of regions and the number of inference steps. Empirically, synthesizing a single 1,024$\times$1,024 image over 50 steps using SDXL basically takes $\sim$ 6 seconds on an NVIDIA A100 GPU.

		Each upscaling stage in {DemoFusion} comprises two denosing paradigms: \textit{regional denosing} and \textit{dilated sampling}. When upscaling an image to ($f$*1,024)$\times$($f$*1,024) resolution ($f$: scale factor), \textit{regional denoising} processes $(2*f-1)^2$ regions and \textit{dilated sampling} processes $f^2$ regions. Each region is 1,024$\times$1,024 and requires 50 inference steps. Therefore, the total theoretical inference cost of DemoFusion to generate a 4,096$\times$4,096 image is $\sum_{f=2}^{4} ((2\times f-1)^2+ f^2)\times6 = 672$ seconds = $11.2$ minutes.

		For upscaling 1,024 to 4,096, {C-Upscale} processes \\$({4,096}/{1,024} \times 2 -1)^2=49$ regions, each requires $50 \times 0.45 = 22$ steps. The total inference cost of regional denoising is $49 \times 0.45 \times 6= 132.3$ seconds. Regarding the three priors utilized in C-Upscale, \textit{GSP} only involves simple tensor transformation and multiplication, which is negligible comparing to cost of UNet inference. \textit{RAP} requires UNet inference on the noisy low-resolution image to obtain the attention prior, equivalent to denoising one region for 22 steps and consumes $2.7$ seconds. \textit{RSP} involves synthesizing $49$ prompts using LLaVA, each takes $2.4$ seconds. Therefore, the total inference time of C-Upscale is $132.3 + 2.7 + 49 \times 2.4=252.6$ seconds $=4.21$ minutes.

		Regarding ScaleCrafter, although the theoretical floating-point operations (FLOPs) are comparable to those of SDXL, the utilization of alternating dilated convolutions requires sparse memory access. This incurs significant time overhead, resulting in increased actual inference time.

		\noindent
	\noindent
\textbf{Acceleration via Distilled Model and Optimized Implementations.} Despite the efficiency gains, upscaling a 1,024$\times$1,024 image to 4,096$\times$4,096 using C-Upscale still requires several minutes. To further enhance computational efficiency, we explore the use of distilled few-step models and optimized implementations. Specifically, we integrate C-Upscale into the pre-trained Hyper-SDXL \cite{hypersd} and utilize vLLM \cite{kwon2023efficient} to accelerate LLaVA inference. These integrations significantly reduce the computational overhead of upscaling a 1,024$\times$1,024 image to 4,096$\times$4,096 from $\sim$ 4 minutes to $\sim$ 30 seconds. The upscaled images produced by the accelerated C-Upscale are presented in Figure \ref{fig:sd}, demonstrating that it consistently maintains high visual quality even with an 8$\times$ increase in inference speed.
	
		\begin{figure}[t]
			\centering
			\includegraphics[width=0.95\columnwidth]{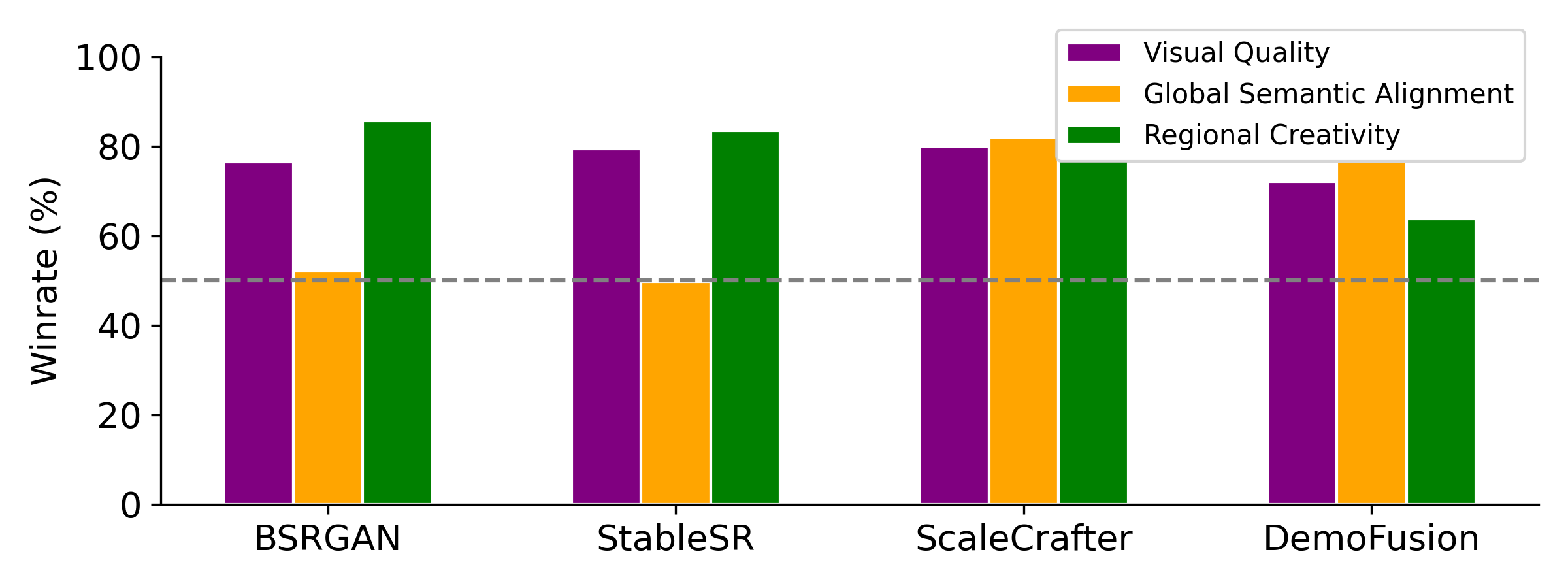}
			\vspace{-0.2cm}
			\caption{
				Human preference ratios of our C-Upscale when compared to each baseline.
			}
			\label{fig:human_eval}
			\vspace{-0.5cm}
		\end{figure}
		
		\subsection{Human Study}\label{sec:human}
		Prior works \cite{dai2023emu,kirstain2024pick,sdxl} have pointed out the weak correlation between existing evaluation metrics and human perception regarding image quality. Thus we additionally conduct human study to examine whether the synthetic higher-resolution images conform to human preferences. Here we quantitatively evaluate C-Upscale against three baselines (i.e., BSRGAN, StableSR, ScaleCrafter, DemoFusion) by comparing each pair for upscaling 100 randomly sampled real-world low-resolution images in LAION-5B. We invite 20 evaluators (10 males and 10 females) with diverse education backgrounds, and ask each evaluator to choose the better one from two synthetic higher-resolution images generated by two different methods given the same low-resolution input, by considering the following three aspects: (1) the fidelity of the visual appearance (\textbf{visual quality}), (2) the semantic alignment to the global text prompt (\textbf{global semantic alignment}), (3) the richness of regional details (\textbf{regional creativity}). Figure \ref{fig:human_eval} illustrates the human preference ratios for each pair of methods. In general, C-Upscale clearly outperforms the two diffusion-based tuning-free methods (ScaleCrafter and DemoFusion) with higher human preference ratios for each aspect. The results demonstrate C-Upscale manages to produce higher-quality images with better semantic alignment and more creative regional details through amplified guidance of global-regional priors. In addition, compared to image super-resolution method of BSRGAN and StableSR, C-Upscale shows comparable performances in global semantic alignment, while superior human preference ratios in visual quality and regional creativity are attained.
		
	\subsection{Discussion on Super-resolution Methods}	
	
			Admittedly, super-resolution approaches do generate new and plausible details to some extent, particularly in cases of severe degradations. Nevertheless, the details generated by SR methods and our C-Upscale exhibit different characteristics. Existing SR models are primarily designed to reconstruct high-resolution images from their low-resolution counterparts, ensuring faithfulness to the input images. These models are trained on synthetic datasets, where low-resolution images are generated by applying degradation processes like downsampling and JPEG compression to high-resolution images. As a result, the details generated by SR methods are inherently shaped and constrained by the specific degradation methods to remove blurriness and JPEG artifacts. In contrast, our C-Upscale leverages diffusion models pre-trained on web-scale datasets, offering greater flexibility in generating creative details based on learned internal knowledge. \\ \indent    
			We also want to emphasize that we fully acknowledge the importance of SR methods, particularly in applications requiring faithful adherence to low-resolution inputs, such as NVIDIA DLSS. Instead, our C-Upscale is more suitable for creative applications, where generating visually appealing and high-resolution images is more important than strictly adhering to the low-resolution input. 
		
		\section{Conclusion}
		
		We have presented C-Upscale, a new tuning-free diffusion-based technique for higher-resolution image generation. C-Upscale excavates three kinds of prior knowledge from low-resolution image, global prompt and estimated regional prompt (i.e., global structure prior, regional attention prior, and regional semantic prior). Such global-regional priors is exploited to strengthen global semantic alignment and regional creativity of output higher-resolution images. C-Upscale can be readily integrated into existing diffusion models and supports image upscaling of both synthetic and real-world inputs. Extensive experiments validate the competitiveness of our C-Upscale when compared to state-of-the-art tuning-free approaches.

        \textbf{Acknowledgments.} This work was supported in part by the Beijing Municipal Science and Technology Project No. Z241100001324002 and Beijing Nova Program No. 20240484681.
		
		\textbf{Data Availability.}
		The synthetic test low-resolution images for upscaling are generated with open-source pre-trained diffusion models, including SD1.5 \cite{ldm}, SDXL \cite{sdxl}, DreamShaper XL \cite{DreamShaper} and Pixart-$\alpha$ \cite{pixart}. The test prompts are randomly sampled from LAION-5B \cite{schuhmann2022laion} and MS-COCO \cite{lin2014microsoft} datasets. The real-world test images are ground-truth images corresponding to 1K sampled LAION-5B prompts.
		
		\bibliographystyle{spmpsci}      
		\bibliography{ref}   
		
	\end{document}